\documentclass{article}
\usepackage{multirow}
\usepackage[table,xcdraw]{xcolor}

\usepackage{ijcai24}

\usepackage{times}
\usepackage{soul}
\usepackage{url}
\usepackage[hidelinks]{hyperref}
\usepackage[utf8]{inputenc}
\usepackage[small]{caption}
\usepackage{graphicx}
\usepackage{amsmath}
\usepackage{amsthm}
\usepackage{booktabs}
\usepackage{algorithm}
\usepackage{algorithmic}
\usepackage[switch]{lineno}
\usepackage{graphicx}

\usepackage{amsfonts}

\newtheorem{theorem}{Theorem}

\title{Converting High-Performance and Low-Latency SNNs through Explicit Modelling of Residual Error in ANNs}

\author{
Zhipeng Huang$^{1*}$
\and
Jianhao Ding$^{2*}$\and
Zhiyu Pan$^2$\and
Haoran Li$^3$\and
Ying Fang$^{1\dag}$\and\\
Zhaofei Yu$^2$\And
Jian K. Liu$^4$\\
\affiliations
$^1$Fujian Normal University\\
$^2$Peking University\\
$^3$Xidian University\\
$^4$University of Birmingham\\
\emails
qsx20221313@student.fjnu.edu.cn
}

\begin{document}

\maketitle

\begin{abstract}
    Spiking neural networks (SNNs) have garnered interest due to their energy efficiency and superior effectiveness on neuromorphic chips compared with traditional artificial neural networks (ANNs). One of the mainstream approaches to implementing deep SNNs is the ANN-SNN conversion, which integrates the efficient training strategy of ANNs with the energy-saving potential and fast inference capability of SNNs. However, under extreme low-latency conditions, the existing conversion theory suggests that the problem of misrepresentation of residual membrane potentials in SNNs, i.e., the inability of IF neurons with a reset-by-subtraction mechanism to respond to residual membrane potentials beyond the range from resting potential to threshold, leads to a performance gap in the converted SNNs compared to the original ANNs. This severely limits the possibility of practical application of SNNs on delay-sensitive edge devices. Existing conversion methods addressing this problem usually involve modifying the state of the conversion spiking neurons. However, these methods do not consider their adaptability and compatibility with neuromorphic chips. We propose a new approach based on explicit modeling of residual errors as additive noise. The noise is incorporated into the activation function of the source ANN, which effectively reduces the residual error. Our experiments on the CIFAR10/100 dataset verify that our approach exceeds the prevailing ANN-SNN conversion methods and directly trained SNNs concerning accuracy and the required time steps. Overall, our method provides new ideas for improving SNN performance under ultra-low-latency conditions and is expected to promote practical neuromorphic hardware applications for further development.
\end{abstract}
\renewcommand{\thefootnote}{}
\makeatletter\def\Hy@Warning#1{}\makeatother
\footnotetext{*these authors contributed equally}
\footnotetext{\dag corresponding author}

\section{Introduction}

\begin{figure}
    \centering
    \includegraphics[width=0.45\textwidth]{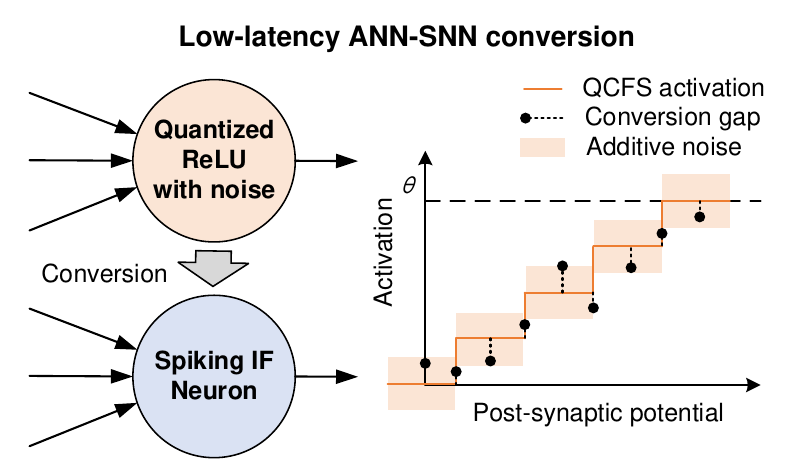}
    \caption{Diagram of our proposed conversion method. State-of-the-art low-latency conversion methods requiring training ANN with quantized activation still bring about a conversion gap in activation. We propose to incorporate additive noise into ANN activation, aiming to compensate for the activation gap.}
    \label{fig:first_fig}
\end{figure}

Spiking neural networks (SNNs), as third-generation artificial neural networks~\cite{maass1997networks} have attracted a lot of attention from researchers, mainly due to their significant advantages deploying on neuromorphic hardware: low power consumption and high efficiency~\cite{farsa2019low,liu2022fpga}. Compared with second-generation artificial neural networks (ANNs) full of float-point multiplication computations, SNNs obtain activation of high sparsity, in which only discrete spikes representing 0 and 1 can do the multiplication-free computation. These attributes enable efficient neuromorphic designs~\cite{pei2019towards,davies2018loihi}, which make SNNs a promising candidate for real-time application~\cite{kim2020spiking,massa2020efficient}.

Nevertheless, non-differentiable spike information transmission makes training high-performance SNNs not a simple task compared with ANN training.
The current major SNN training methods include supervised surrogate backpropagation~\cite{wu2019direct,zenke2021remarkable}, ANN-SNN conversion~\cite{cao2015spiking,diehl2015fast,rueckauer2017conversion}, unsupervised spike-timing-dependent plasticity learning~\cite{kheradpisheh2018stdp,diehl2015unsupervised} as well as some hybrid or online training methods~\cite{lee2018training,tavanaei2019bp,rathi2020enabling,xiao2022online}. 
For deep SNNs, unsupervised learning methods perform poorly on large-scale datasets. Supervised surrogate backpropagation requires expansion along the temporal axis utilizing surrogate functions~\cite{zenke2021remarkable}, which requires huge GPU resources. The current progress of online training methods breaks the temporal dependence between spikes of  neighboring  time steps, yet requires inference multiple time steps and still performs worse among supervised methods. Therefore, in the supervised training domain, ANN-SNN conversion bypasses the non-differentiable problem by training an ANN counterpart and is still promising for large-scale networks and datasets. The key concentration is on reducing the latency that the network needs. For traditional conversion methods, a long inference time is required to match the firing rate of the SNN with the activation value of the ANN~\cite{sengupta2019going}. However, this increases the overhead on neuromorphic chips. Addressing the problem of conversion errors at low latency has become an important challenge in current research. Attempts to lower latency by proposing hybrid methods mixing conversion and surrogate training still fail to decrease the training cost.

The challenge of low-latency ANN-SNN conversion arises from conversion errors, which have been identified by previous studies~\cite{wang2205towards,bu2023optimal}, resulting in a performance gap under low-latency conditions. To eliminate these errors, Bu et al.~\shortcite{bu2023optimal} employed quantized QCFS activation in the source ANN from the assumption of bounded residual membrane potentials. They didn't explicitly optimize the so-called ``unevenness error'' in their work, thus the performance of QCFS is acceptable at 8-time steps but poor at ultra-low time steps (e.g., T $\leq$ 4). Efforts to further eliminate the conversion errors with low latency include shifting the initial membrane potential of spiking neurons on the fly~\cite{hao2023bridging} and BPTT fine-tuning after conversion~\cite{wang2205towards}. These methods flaw as they either need modifications to neuromorphic hardware or run extra surrogate training on GPUs. By contrast, eliminating conversion errors in ANN training is ideal as it introduces no extra cost. Thus, we propose to further explicitly model residual conversion errors and incorporate them into ANN training.
The main contributions of this paper are summarised below:
\begin{itemize}
    \item We find that the conversion loss for low-latency SNN primarily stems from residual errors between quantized ANNs and converted SNNs. We experimentally observed the distribution of residual errors in each layer with low time steps and find that the variance of the residual error is significant while the mean is rather small.
    \item Based on the observation, we propose to explicitly model the residual error as a Gaussian noise with a zero mean and integrate the noise into the quantized activation of the source ANN during training, aiming to compensate for the gap between the source ANN and the converted SNN.
    \item Due to the difficulty of predicting and estimating residual errors after conversion, we induct the noise intensity from a small validation set and apply a layer-wise error-compensating strategy while training.
    \item We demonstrate the effectiveness of our method on CIFAR10/100. Under low-latency conditions, our method outperforms previous state-of-the-art methods and shows improvement in performance with low time steps. For example, we attain an outstanding top-1 accuracy of 93.72\% on CIFAR-10 with just 2 time steps.   

\end{itemize}

\section{Related Work}


The core idea of ANN-SNN conversion is to combine the computational efficiency of ANNs and the biological rationality of SNNs to accomplish specific AI tasks. This approach avoids the huge computational consumption associated with directly training SNNs~\cite{zhang2020temporal} and is a feasible solution for training deep SNNs. Cao et al.~\shortcite{cao2015spiking} found the equivalence between ReLU activation and the fire rate of the IF neuron, laying the foundation for ANN-SNN conversion based on spike firing rate. 
Subsequently, various weight-threshold balance methods have been popular in reducing the inference delay (denoted as $T$) to typically up to hundreds~\cite{diehl2015fast,rueckauer2017conversion,sengupta2019going,han2020rmp,ho2021tcl}, yet the delay is still unacceptable for large-scale architectures.
To further improve the performance, some works have introduced methods to modify neurons, such as burst spikes~\cite{li2022efficient} and symbolic signed neurons~\cite{wang2022signed,li2022quantization}. However, they destroy the binary property of spiking neurons, making it difficult to apply to neuromorphic chips. 

Recent advances in ANN-SNN conversion call for low-latency conversion, where conversion error reduction becomes the focus of research. The conversion error is the gap between the ANN activation and the firing rate of SNN.
Deng et al.~\shortcite{deng2021optimal} proposed ThreshRelu and found that the conversion error can be optimized layer by layer. Later, quantization clip-floor-shift (QCFS) activation proposed by \cite{bu2023optimal} simulates the characteristics of SNNs as much as possible and has already achieved good performance under low-latency conditions ($T\leq$ 8). However, there is still a performance gap between ANN and SNN when the latency is lower. The fundamental reason is that the so-called "unevenness error" pointed out by the authors has not been effectively solved. This is precisely an important challenge. To resolve this error, Hao et al. proposed SRP~\cite{hao2023reducing} and OffsetSpikes methods~\cite{hao2023bridging}. However, these methods include restricting the converted SNN to release and move the initial membrane potential, increasing the overhead in deployment. Wang et al.~\shortcite{wang2205towards} considered modifying the initial membrane potentials and fine-tuning them by surrogate training in the converted SNN to solve the unevenness error. While these methods improve performance under low latency compared to previous methods, they do not consider the usability of the methods in neuromorphic hardware chips. The goal of ANN-SNN conversion is to take advantage of GPU training of ANNs to create a well-trained, low-latency SNN model without introducing complex operations. The paper aims to further examine the conversion error and reduce the performance gap between ANNs and SNNs at low latency by modeling the error in the activation function of the source ANN. This approach will help improve the performance of SNNs on neuromorphic hardware chips.

\section{Preliminaries}

\subsection{Neuron Model}
\textbf{ANN neurons.} For traditional ANNs, the input tensor is fed into the ANN and processed layer by layer through weighted summation and the continuous nonlinear activation function $f(\cdot)$ to produce output activation. The forward propagation for neurons in layer $l$ in an ANN can be expressed as:
\begin{equation}
    \boldsymbol{a}^{l}=f\left(\boldsymbol{W}^{l} \boldsymbol{a}^{l-1}\right), \quad l=1,2, \ldots, M
    \label{eq:ann_act}
\end{equation}%
where $\boldsymbol a^{l-1}$ and $\boldsymbol{a}^{l}$ are the pre-activation and post-activation vectors of the $l$-th layer, $\boldsymbol{W}^l$ is the weight matrix, and $f(\cdot)$ is usually set to the ReLU activation function.


\textbf{SNN neurons.} Unlike ANNs, SNNs introduce a temporal dimension and employ non-differentiable spike activation. For deep SNNs, the inputs are usually repeated over $T$ time steps during forward propagation for the purpose of better performance, which can produce the final mean output as logits~\cite{zheng2021going,xu2023constructing}. We deployed the integrate-and-fire (IF) neuron model~\cite{gerstner2002spiking} as reported in previous studies of ANN-SNN conversion~\cite{cao2015spiking,diehl2015fast}. The overall discrete kinetics of the IF neuron can be expressed as follows:
\begin{align}
\label{eq:reset}
    \boldsymbol{v}^{l}(t) &=\boldsymbol{v}^{l}(t-1)+\boldsymbol{I}^{l}(t)-\boldsymbol{s}^{l}(t-1) \theta^{l},\\ 
    \boldsymbol{I}^{l}(t) &=\boldsymbol{W}^{l} \boldsymbol{s}^{l-1}(t) \theta^{l-1}=\boldsymbol{W}^{l}\boldsymbol x^{l-1}(t). 
\end{align}%

Here, $\boldsymbol v^l(t)$ and $\boldsymbol{I}^{l}(t)$ denote the membrane potential and the input current of layer $l$ at time step $t$, respectively. $\boldsymbol W^l$ is the synaptic weight, and $\theta^l$ is the firing threshold. $\boldsymbol s^l (t)$ indicates whether the spike is triggered at time step $t$. For the $i$th neuron, the neuron fires a spike if the potential after charging $u_i^l (t) ={v}_i^{l}(t-1)+I_i^{l}(t)$ exceeds the firing threshold $\theta^l$. The subscript $i$ denotes the $i$th element of the vector unless otherwise specified. This firing rule can be described by the following equation:
\begin{equation}
    \resizebox{.85\linewidth}{!}{$
            s_{i}^{l}(t)=H\left(u_{i}^{l}(t)-\theta^{l}\right)=\left\{\begin{array}{ll}1, & u_{i}^{l}(t) \geq \theta^{l} \\0, & u_{i}^{l}(t)<\theta^{l}\end{array}\right.
        $}
\end{equation}
where $H(\cdot)$ is the Heaviside step function. 
$\boldsymbol x^{l-1}(t)=\boldsymbol{s}^{l-1}(t) \theta^{l-1}$ denotes the post-synaptic potential of neurons in layer $l-1$ as introduced by Bu et al.~\shortcite{bu2023optimal}. In addition, to minimize information loss during inference, neurons in SNN employ a reset-by-subtraction mechanism in Eq.~\ref{eq:reset}, that is, once a spike is triggered, the membrane potential after the spike needs to be subtracted by the firing threshold $\theta^l$.
 
\subsection{ANN-SNN Conversion}

The idea behind ANN-SNN conversion is to create a mapping from spiking neurons' postsynaptic potential in SNN to ReLU activation in ANN. Denote that $\boldsymbol \phi^{l}(T)=\frac{\sum_{t=1}^T \boldsymbol x^{l}(t)}{T}$ is the average post-synaptic potential during $T$ time steps. The average post-synaptic potential of neurons in the neighboring layers is related as follows:
\begin{equation}
    \boldsymbol \phi^l(T)=\boldsymbol{W}^l \boldsymbol \phi^{l-1}(T)-\frac{\boldsymbol{v}^l(T)-\boldsymbol{v}^l(0)}{T}. \label{eq:postsynaptic_relate}
\end{equation}
The derivation of Eq.~\ref{eq:postsynaptic_relate} can be found in the work of Bu et al.~\shortcite{bu2023optimal}. It is obvious that  $\boldsymbol\phi^{l}(T)$ is in the range of $[0,\theta^{l}]$ and only takes discrete values due to $\boldsymbol x^{l}(t)=\boldsymbol{s}^{l}(t) \theta^{l}$. Since $\phi^l(T)>0$, one can build the equivalent mapping from ANN activation $\boldsymbol{a}^l$ to $\boldsymbol \phi^{l}(T)$ by constraining $\boldsymbol{a}^l$ into discrete values and setting a specific upper bound for $\boldsymbol{a}^l$. If $T\rightarrow\infty$, $(\boldsymbol{v}^l(0) - \boldsymbol{v}^l(T))/T\rightarrow0$, in this case, lossless ANN-SNN conversion can be achieved. Nevertheless, a large value of $T$ significantly increases the cost of the application of efficient SNN hardware, which calls for low-latency conversion. When $T$ is small, $(\boldsymbol{v}^l(0) - \boldsymbol{v}^l(T))/T$ is not approaching 0, which indicates the existence of conversion errors.

To address the low-latency conversion, Bu et al.~\shortcite{bu2023optimal} proposed a concept based on the finite residual membrane potential assumption $\boldsymbol v(T) \in [0,\theta^l]$ and 
derived the QCFS activation function as a substitute for the commonly used ReLU activation function in source ANNs: %
\begin{equation}
\boldsymbol{a}^l=\text{QCFS}\left( \boldsymbol{z}^l \right) =\lambda ^l\mathrm{clip}\left( \left. \frac{1}{L}\left. \left\lfloor \frac{\boldsymbol{z}^lL}{\lambda ^l}+0.5 \right. \right. \right\rfloor ,0,1 \right) ,
\label{eq:QCFS}
\end{equation}
where $L$ indicates the activation quantization step, $\boldsymbol{z}^l=\boldsymbol{W}^l\boldsymbol{a}^{l-1}$. $\lambda^l$ is the trainable threshold for layer $l$ in ANN. When one performs a conversion,  $\lambda^l$ should be copied as $\theta^l$ in SNN. 
Such that given the same average input $\boldsymbol{z}^l$ as ANN, when $T=L$,  $\boldsymbol{v}^l(0)$ is set to $0.5\cdot \theta_l$, and $\boldsymbol{v}^l(T) \in [0,\theta^l]$, the converted average postsynaptic potential can be expressed as:
\begin{equation}
    \boldsymbol{\phi}'^l (T)=\theta ^l\mathrm{clip}\left( \left. \frac{1}{T}\left. \left\lfloor \frac{\boldsymbol{z}^lT+\boldsymbol{v}^l(0)}{\theta ^l} \right. \right. \right\rfloor ,0,1 \right).
    \label{eq:QCFS_cvt}
\end{equation}
It is proven that such a source ANN activation function can more accurately approximate the SNN's activation function and eliminate quantization error in expectation. Other work, such as~\cite{li2022quantization,jiang2023unified} also shared a similar quantization framework. After applying quantization to ANN, the low-latency performance has significantly improved. However, this work still incurs unresolved conversion errors, which we will brief in the following section.



\begin{figure}
    \centering
    \includegraphics[scale=0.40]{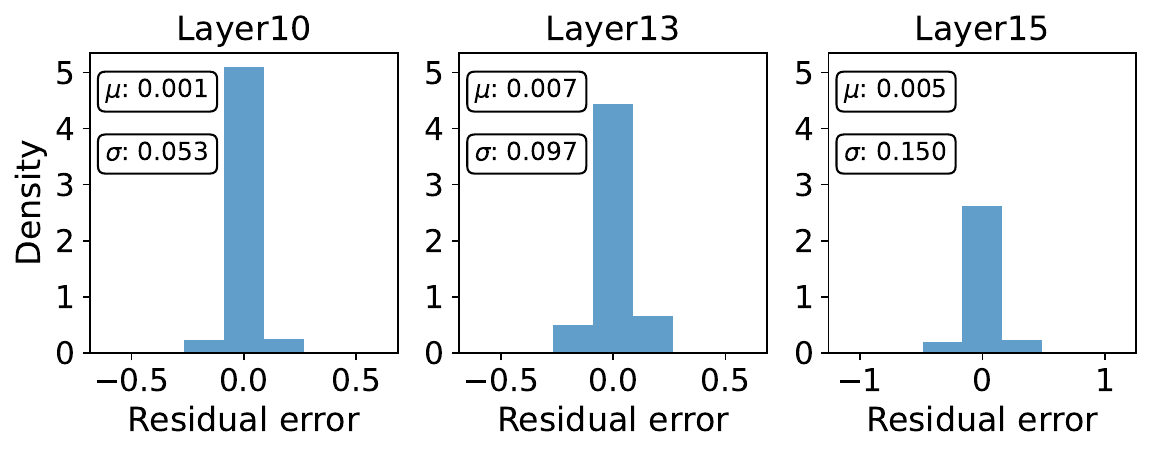}
    \caption{The distribution of the residual error of ANN output and the average postsynaptic potential for SNN. We calculate the residual error of some activation layers during the training of VGG16 on the CIFAR10 dataset.}
    \label{fig:gap distribution}
\end{figure}

\section{Residual Error in Low-Latency Conversion}
\label{sec:Error Analysis}

In this section, we discuss three potential errors that occurred in the process of converting ANN to SNN, i.e., clipping error, quantization error, and residual error, which contribute to the performance gap between source ANNs and target SNNs. 
Moreover, we examine the distribution of the residual error.

\subsection{Clipping and Quantization error}

Clipping and quantization errors bear a certain resemblance as they are both due to the difference between $a^{l}$ and $\phi^{l}(T)$, where $a^{l}$ is the ANN quantized activation of layer $l$ and $\phi^{l}(T)$ is the average postsynaptic potential from layer $l$ in converted SNN.
Clipping error arises from the difference in value ranges. In Eq.~\ref{eq:QCFS}, if $\hat{\lambda}^l$ is the actual maximum value of output $a^l$ and larger than $\theta^{l}$, then the unbounded activation larger than $\theta^l$ cannot be exactly expressed after conversion. This may cause clipping errors.

Quantization error arises from differences in distribution. Output $a^{l}$ in ANN is continuous values spread all over the range of $[0,a^{l}_{max}]$, while $\phi^{l}(T)$ can be distributed only on discrete values like $\frac{\theta^l}{T},\frac{2\theta^l}{T},\frac{3\theta^l}{T},\cdots$, thus they cannot be perfectly matched. 

Reducing clipping and quantization errors can be achieved by modifying the activation functions of the source ANN to quantized ones~\cite{li2021free}. Besides, adding the trainable thresholds is proven to be effective as it can reduce the clipping error to zero by directly mapping the trainable upper bound of the activation to the threshold of the SNN~\cite{ho2021tcl}.

\subsection{Residual error}  
Previous studies have noticed one error beyond clipping and quantization errors, which usually attribute to neuronal differences. These errors are used to considered as transient dynamics~\cite{rueckauer2017conversion}, temporal jitter of spike trains~\cite{wu2021tandem}, or unevenness error~\cite{bu2023optimal}. We deem that these studies do not accurately reflect the cause of the error. Here, we refer to this type of error as residual error. Firstly, the primary reason leading to this error is that IF neurons with reset-by-subtraction mechanisms fail to respond to residual membrane potentials outside the range of $[0, \theta]$. Besides, when the quantization parameter $L$ mismatches the inference time step $T$, the average postsynaptic membrane potential also mismatches the activation.

Residual error seriously affects the performance of SNNs under low-latency conditions. Here, we would like to give the form of residual error denoted as $g^l(T)$ for layer $l$. IF neurons in layer $l$ receive weighted input $\boldsymbol{\hat{z}}^l=\boldsymbol{W}^l \boldsymbol\phi ^{l-1}(T)$, and their initial membrane potential is denoted as $\boldsymbol{v}^l(0)$. Then we can reformulate Eq.~\ref{eq:postsynaptic_relate} into:
\begin{equation}
   \boldsymbol\phi^{l}(T)=\theta^{l} \operatorname{clip}\left(\frac{1}{T}\left\lfloor\frac{\boldsymbol{\hat{z}}^l T+\boldsymbol{v}^{l}(0)}{\theta^{l}}\right\rfloor, 0,1\right) + \boldsymbol g^l(T).
   \label{eq:An exact expression for the mean post-synaptic potential}
\end{equation}
Eq.~\ref{eq:An exact expression for the mean post-synaptic potential} now is not related to the source ANN and $\boldsymbol g^l(T)$ is not properly modeled. For QCFS conversion, when $T=L$,  $\boldsymbol{v}^l(0)=0.5\cdot \theta_l$, and $\boldsymbol{v}^l(T) \in [0,\theta^l]$, $\boldsymbol{a}^l$ in Eq.~\ref{eq:QCFS} matches $\boldsymbol{\phi }'^l(T)$ in Eq.~\ref{eq:QCFS_cvt}. And we have $\boldsymbol{\phi }^l(T)=\boldsymbol{\phi }'^l(T)+g^l(T)$. 
However, these ideal conditions are too harsh; in fact, we can only obtain:
\begin{align}
    \boldsymbol{\phi }^l(T)&=\boldsymbol{\phi }'^l(T)+\boldsymbol g'^l(T),\label{eq:phi_and_g}\\
    &=\boldsymbol{a}^l+\boldsymbol g'^l(T), \label{eq:a_and_phi}
\end{align}
where $\boldsymbol g'^l(T)$ absorbs the original residual error $\boldsymbol g^l(T)$ and other errors caused by $\boldsymbol{v}^l(T) \notin [0,\theta^l]$ or $T\neq L$.
According to Eq.~\ref{eq:a_and_phi}, we can observe the distribution of $\boldsymbol g'^l(T)$ based on the source ANN activation and converted SNN postsynaptic potential. Specifically, we train an ANN with QCFS activation on CIFAR-100 for VGG-16, and convert it to an SNN by fixing $T=L$. Please refer to Fig.~\ref{fig:gap distribution}.
We find that the standard deviation of the distribution of $\boldsymbol g'^l(T)$ is large while the mean is comparably small (close to 0). The distribution is almost symmetric around 0. This suggests that the distribution of $\boldsymbol g'^l(T)$ will be highly dispersed around 0.

\begin{figure}[t]
  \begin{minipage}[t]{0.5\linewidth}
    \centering
    \includegraphics[scale=0.22]{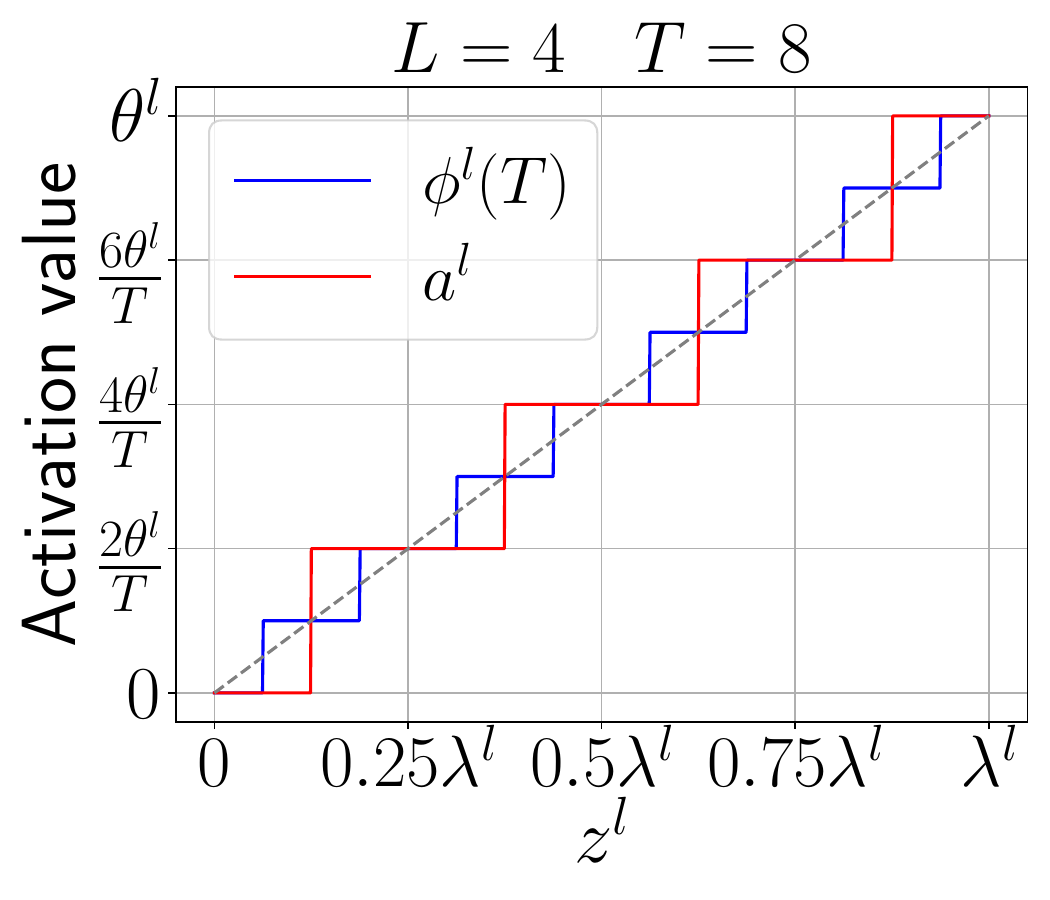}
    \label{fig:QCFS}
  \end{minipage}%
  \begin{minipage}[t]{0.5\linewidth}
    \centering
    \includegraphics[scale=0.22]{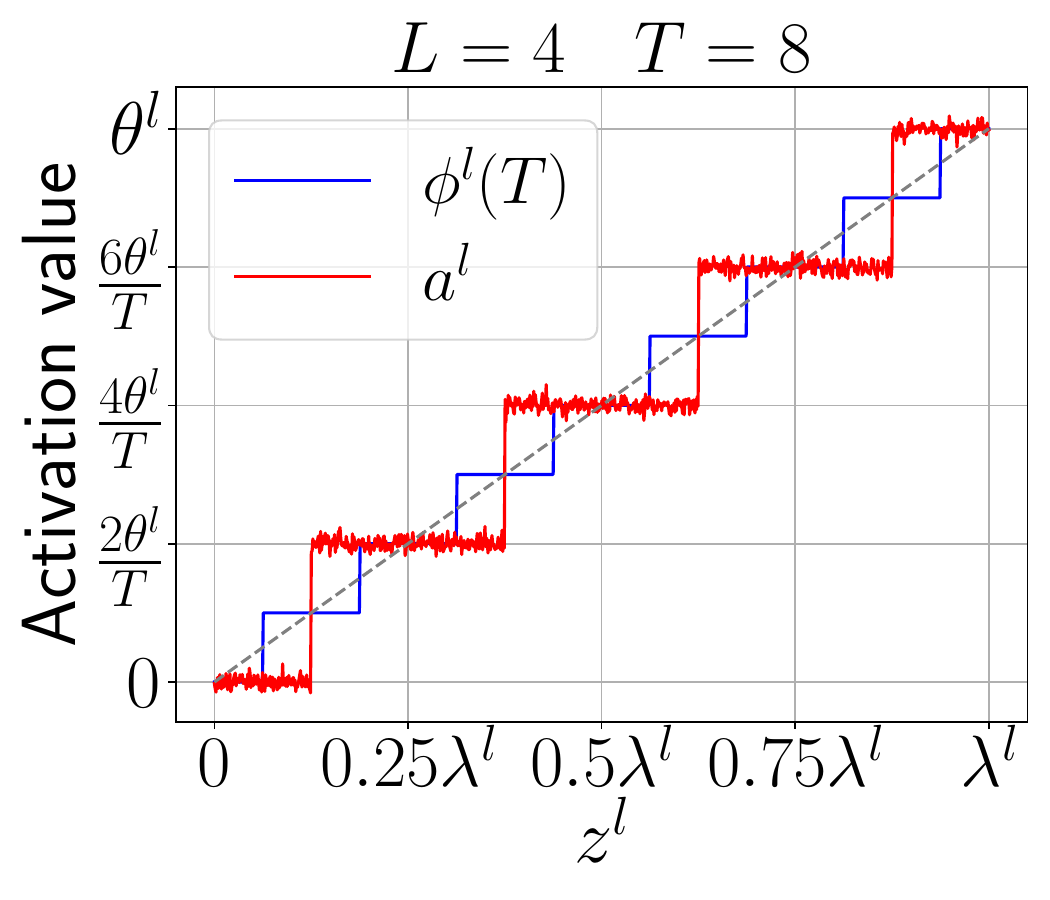}
  \end{minipage}
  \caption{ Comparison of SNN output $\phi^l(T)$ and ANN output $\mathbf{a}^l$ with the same input $\mathbf{z}^l$. The figure shows two activation functions for source ANNs: quantization clip-floor-shift (QCFS) activation (left) and our proposed Noisy Quantized activation with residual error noise modeling (right).}
  \label{fig:our Activation}
\end{figure}

\section{Methods}
  In this section, we come up with an explicit modeling method and improve activation functions by adding Gaussian noise to lessen the error between ANN and converted SNN. 
  We manage to improve the low-latency conversion performance by adding a fixed noise for all layers.
  Finally, we propose a layer-wise error-compensating strategy to set the noise intensity more accurately for each activation layer.
\subsection{Explicit Modeling of Residual Error}

Since the distribution of $\boldsymbol g'^l(T)$ is almost symmetric around 0 and has a significant standard deviation, we consider using $\delta^l \cdot \boldsymbol{G}$ as an approximate alternative to $\boldsymbol g'^l(T)$, where $G_i$ is the $i$th item in $\boldsymbol{G}$, $G_i  \sim N(0,1)$, and $\delta^l \cdot G_i $ is sampled from an i.i.d. Gaussian distribution for each neuron $i$ in layer $l$.
\begin{equation}
   g'^l_i(T) \approx \delta^l \cdot G_i,
    \label{eq:The gaussion estimate for g^l(T)}
\end{equation}
where $g'^l_i(T)$ is the $i$th item in $\boldsymbol g'^l(T)$. In this case, $\delta^l \cdot G_i  \sim N(0,{\delta^l}^2)$ for each neuron $i$.
With such an approximate, by combining Eq.~\ref{eq:phi_and_g} and Eq.~\ref{eq:The gaussion estimate for g^l(T)}, we update a more accurate estimation expression for $\mathbf{\phi}^l$ in the SNN:%
\begin{equation}
    \phi^{l}(T) \approx \boldsymbol{\phi }'^l(T)+ \delta^l \cdot \boldsymbol{G}.
    \label{eq:An expression for the estimation of post-synaptic mean potentials}
\end{equation}
We consider the additional Gaussian noise as compensation for the residual error. 
According to Eq.~\ref{eq:phi_and_g}, we propose to introduce an quantized activation function with a parameterized noise model of the residual error to train the ANN:%
\begin{equation}
    \boldsymbol{a}^l=\text{NQ}\left(\boldsymbol{z}^l\right)=\lambda^l \operatorname{clip}\left(\frac{1}{L}\left\lfloor\frac{\boldsymbol{z}^l L}{\lambda^l}+0.5\right\rfloor, 0,1\right)+\delta^l \cdot \boldsymbol G
    \label{eq:our activation function}
\end{equation}
We name this activation Noisy Quantized activation (NQ). In Eq.~\ref{eq:our activation function}, considering that the effect of quantization error cannot be offset in the case of $T\ne L$, we adopt a treatment similar to that of ~\cite{bu2023optimal}, i.e., introducing a shift term of 0.5 in the activation function. Note that Eq.~\ref{eq:our activation function} degenerates to a QCFS activation function when $\delta^l=0$.

Based on the NQ activation and the average postsynaptic potential of the converted SNN activation in Eq.~\ref{eq:postsynaptic_relate}, we can derive the conversion error $\boldsymbol \epsilon^{l}$ between the ANN and the SNN:
\begin{equation}
    \boldsymbol \epsilon^{l} = \boldsymbol{W}^l \boldsymbol \phi^{l-1}(T)-\frac{\boldsymbol{v}^l(T)-\boldsymbol{v}^l(0)}{T} - \text{NQ}(\boldsymbol{z}^l).
    \label{eq:Estimates of absolute error between ANN and SNN}
\end{equation}
With the definition of conversion error above, we show Theorem~\ref{theorem:1}, which proves that under some conditions, the expectation of the conversion error is zero.
\begin{theorem}
\label{theorem:1}
Given an ANN using our proposed NQ activation in Eq.~\ref{eq:our activation function}, the trained ANN is converted to an SNN with IF neurons with the same weights and satisfying $\theta^l = \lambda^l$ for each layer. Assume that $v^l(0)=\frac{\theta^l}{2}$, $\boldsymbol{v}^l(T) \in [0,\theta^l]$ where $\boldsymbol{v}^l(T)$ is the membrane potential at time $T$. Then we have:
\begin{equation}
    \forall T, L,\delta^l \quad \mathbb{E}_{z}(\widetilde{\epsilon})=\mathbf{0} .
\end{equation}%
\end{theorem}
The proof of Theorem~\ref{theorem:1} given in the Appendix. Theorem~\ref{theorem:1} shows that for any $\delta^l$, even if $L \ne T$, the additional introduction of $\delta^l \boldsymbol G$ in the NQ activation function modeled is complementary to the quantization and clipping error. $\delta^l \boldsymbol G$ does not affect the expected value of the conversion error. These nice properties ensure the feasibility of our high-performance converted SNN at ultra-low inference time steps.

\begin{figure}
  \begin{minipage}[t]{0.5\linewidth}
    \centering
    \includegraphics[scale=0.24]{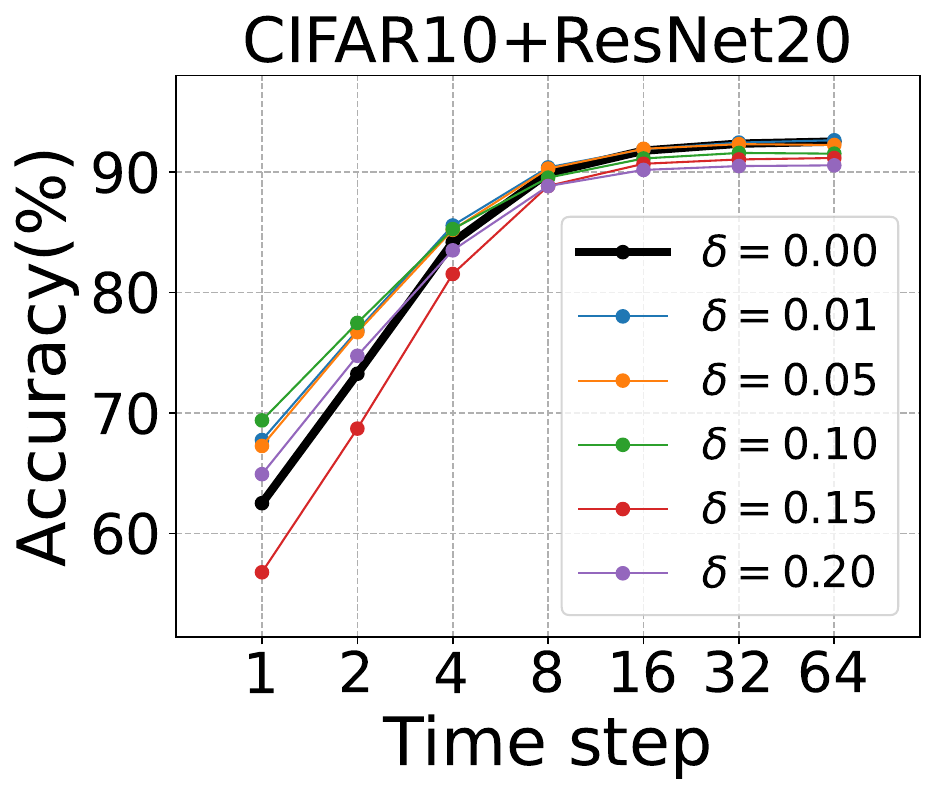}
  \end{minipage}%
  \begin{minipage}[t]{0.5\linewidth}
    \centering
    \includegraphics[scale=0.24]{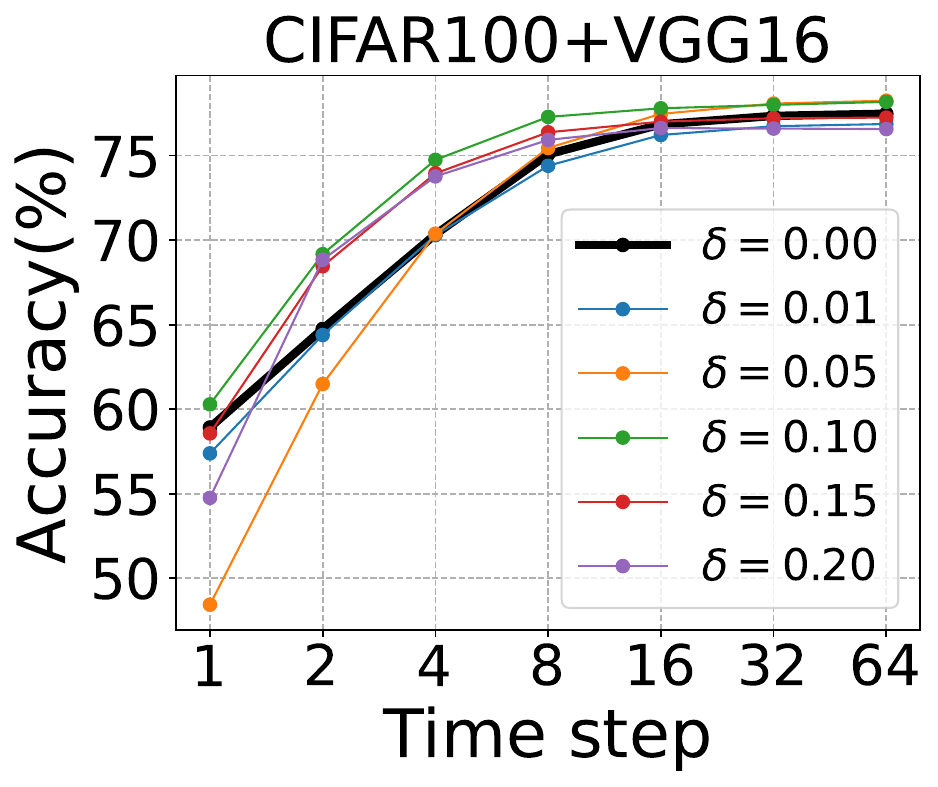}
  \end{minipage}
  \caption{The effect of noise intensity. Adding a certain amount of noise to the activation function benefits inference performance at short time steps.}
  \label{fig:The effect of noise intensity}
\end{figure}

\subsection{Minimizing the Gap Between ANN and SNN}

Determining the appropriate Gaussian noise intensity $\delta^l$ in real training is a critical task. Initially, we consider manually setting a fixed noise intensity for all activation layers. To evaluate the effect of adding noise to the activation on the performance of ANN-SNN conversion, we conduct experiments for VGG16 on the CIFAR10 dataset and ResNet20 on the CIFAR100 dataset. The experimental results are shown in Fig.~\ref{fig:The effect of noise intensity}. Specifically, we add a fixed and shared zero-mean Gaussian distribution to all activation layers during the training of the ANN, where $\delta^l = \delta$ for each layer. We try different settings of the noise intensity $\delta$. The experimental results in Fig.~\ref{fig:The effect of noise intensity} show that different settings of the noise intensity will directly affect the conversion performance. In the case of adding less noise, the performance of the converted SNN is improved for all time steps, but there is still room for further improvement. However, when the noise intensity is too large, we observe that the converted SNN fails to converge to a higher accuracy. We believe that this is due to the fact that the noise introduces randomness that affects the accuracy of the ANN and further affects the accuracy of the SNN. To obtain a fixed noise setting that is most beneficial for ANN-SNN conversion usually requires a lot of experimentation and tuning for different datasets and network structures. In addition, according to our observation on residual error across different layers in Fig.~\ref{fig:gap distribution}, it can be seen that the error distributions across layers are not consistent. Therefore, personalizing the noise setting for each activation layer would be a more reasonable way.

We propose a hierarchical error compensation strategy to optimize the ANN-SNN conversion during ANN training. Since it is difficult to predict and estimate the residual error before conversion, we infer the converted SNN on a small validation set and decide the noise intensity for each layer. Specifically, before training, we split a small portion of the training dataset as the validation dataset. For the first epoch, we choose to train using NQ by setting $\delta^l=0$ for all $l$. After training one epoch of the training set, we successively process the validation set with the training ANN with NQ activation and the corresponding SNN. During the validation process, we first calculate the output of each activation layer of the ANN and then the average postsynaptic potential of the corresponding SNN neuron layer at $T = \tau$. After obtaining vectors from ANN and SNN, we calculate the standard deviation of the difference between the two vectors. 
Subsequently, we set the noise intensity $\delta^l$ of each activation layer to the standard deviation obtained, respectively. That is, in each epoch, we adjust the noise intensity $\delta^l$ according to the statistics obtained from the previous epoch on the validation set. Such a procedure ensures that the additive noise is independent within the different activation layers, allowing a more precise compensation of the effects caused by residual error.

\begin{figure}[t]
  \begin{minipage}[t]{0.5\linewidth}
    \centering 
    \includegraphics[scale=0.22,trim={0 0 0 0},clip]{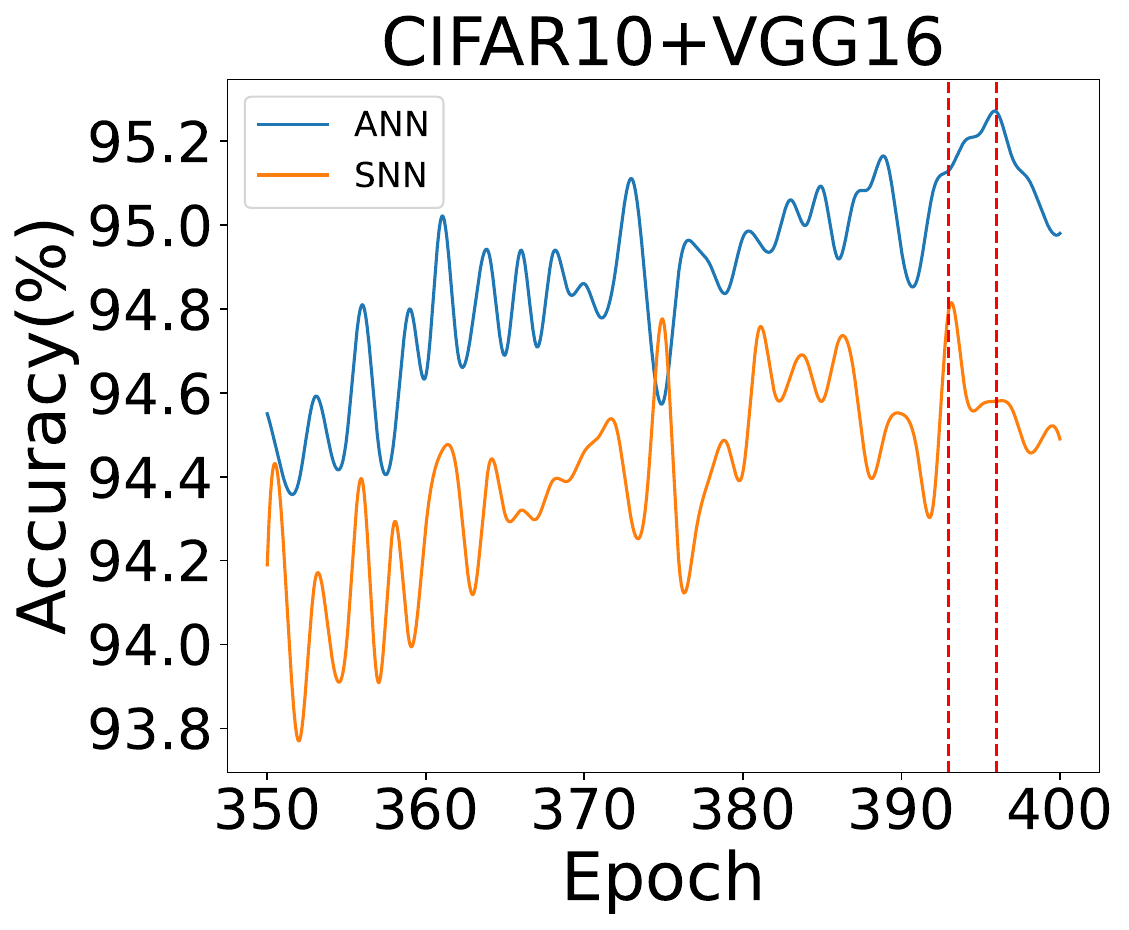}
  \end{minipage}%
  \begin{minipage}[t]{0.5\linewidth}
    \centering
    \includegraphics[scale=0.22,trim={0 0 0 0},clip]{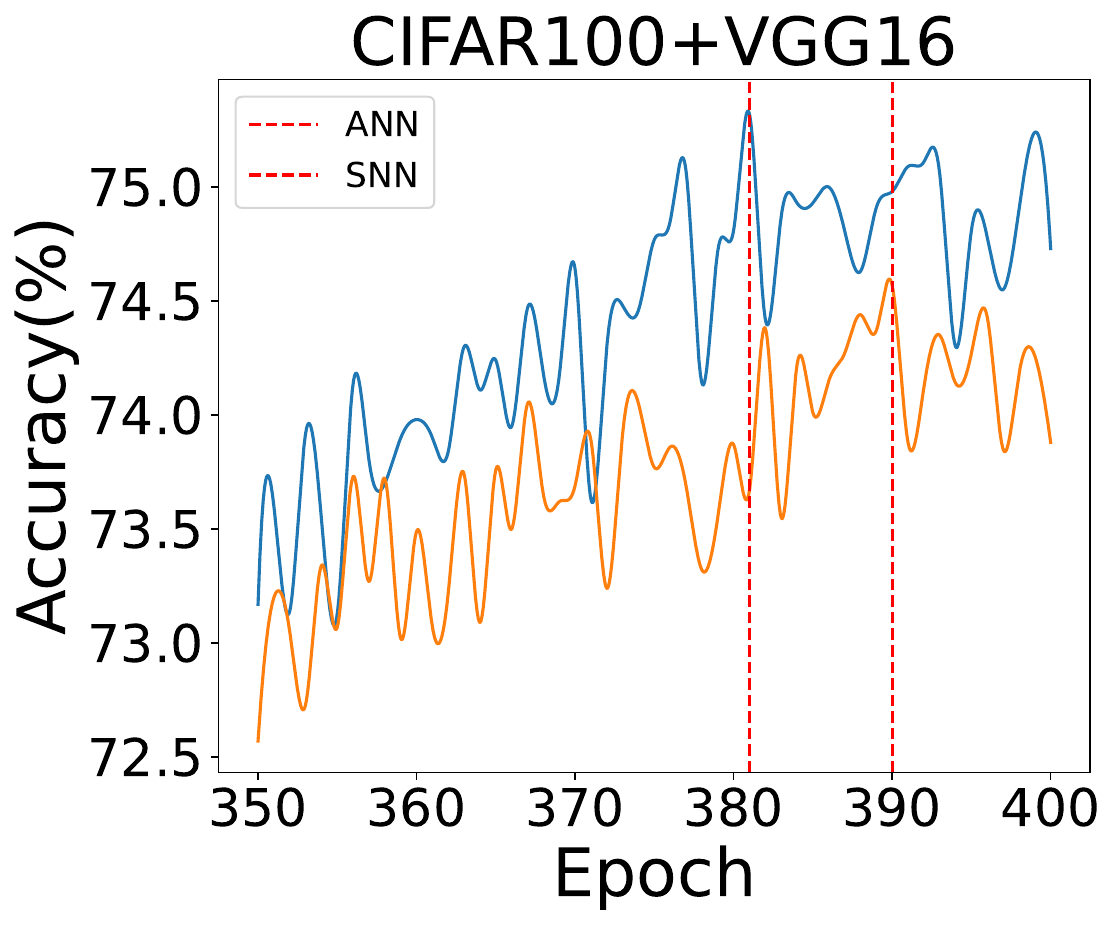}
  \end{minipage}
  \caption{Accuracy curves of ANN and SNN on the testing dataset during the training process}
  \label{fig:Accuracy curves}
\end{figure}

In addition, we find that the ANN accuracy and SNN accuracy on the testing set are not always optimal at the same epoch during the training process due to the introduction of the additive noise. We record the accuracy of VGG16 architecture of ANN and ANN on the CIFAR10 dataset and the CIFAR100 dataset during training and display the accuracy curve in Fig.~\ref{fig:Accuracy curves}. As can be seen from the results in Fig.~\ref{fig:Accuracy curves}, if we save the model with the best ANN performance during training, we will miss the best epoch when SNN have the best performance. Therefore, during training, we choose to evaluate the converted SNN model with $T$ time steps rather than the model that is optimal for the ANN. 

\section{Experiments}
In this section, we use the image classification task to evaluate the effectiveness and performance of our proposed method on the CIFAR-10 and CIFAR-100 datasets. 
We use the widely adopted VGG-16, ResNet-18, and ResNet-20 network structures as source ANNs for conversion.

\subsection{Experimental Setting}

For ANN training, we use the SGD training optimizer and a cosine decay scheduler to tune the learning rate. The initial learning rate is set to 0.1 for CIFAR-10 and 0.05 for CIFAR-100. Besides, we set the weight decay to $5 \times 10^{-4}$ for both datasets. We train all models for 400 epochs. Our setting of the quantization step $L$ is consistent with Bu et al.~\shortcite{bu2023optimal}. When training on the CIFAR-10 dataset, $L$ is set to 4. When training VGG-16 and ResNet-18 on the CIFAR-100 dataset, $L$ is set to 4. When training ResNet-20 on the CIFAR-100 dataset, $L$ is set to 8. When applying the layer-wise strategy, we suggest that the noise-induction time step $\tau$ be consistent with $L$. Please refer to the Appendix for a detailed experimental setting. 

\begin{figure}[t]
    \centering
    \includegraphics[scale=0.45]{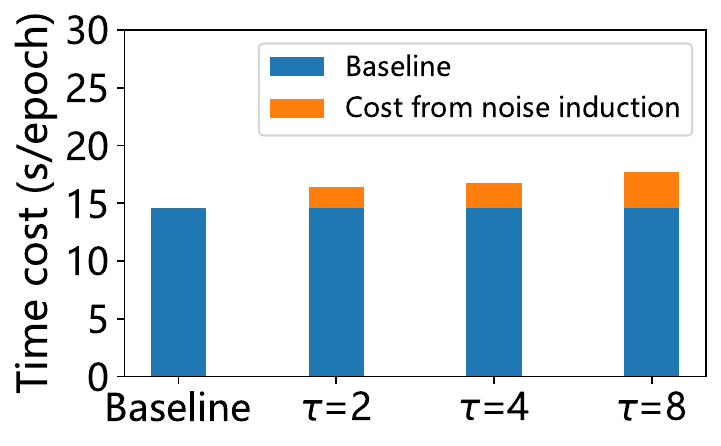}
    \caption{Comparison of training overhead of ANN}
    \label{fig:training_time}
\end{figure}

\begin{figure}[t]
  \begin{minipage}[t]{0.5\linewidth}
    \centering
    \includegraphics[scale=0.24]{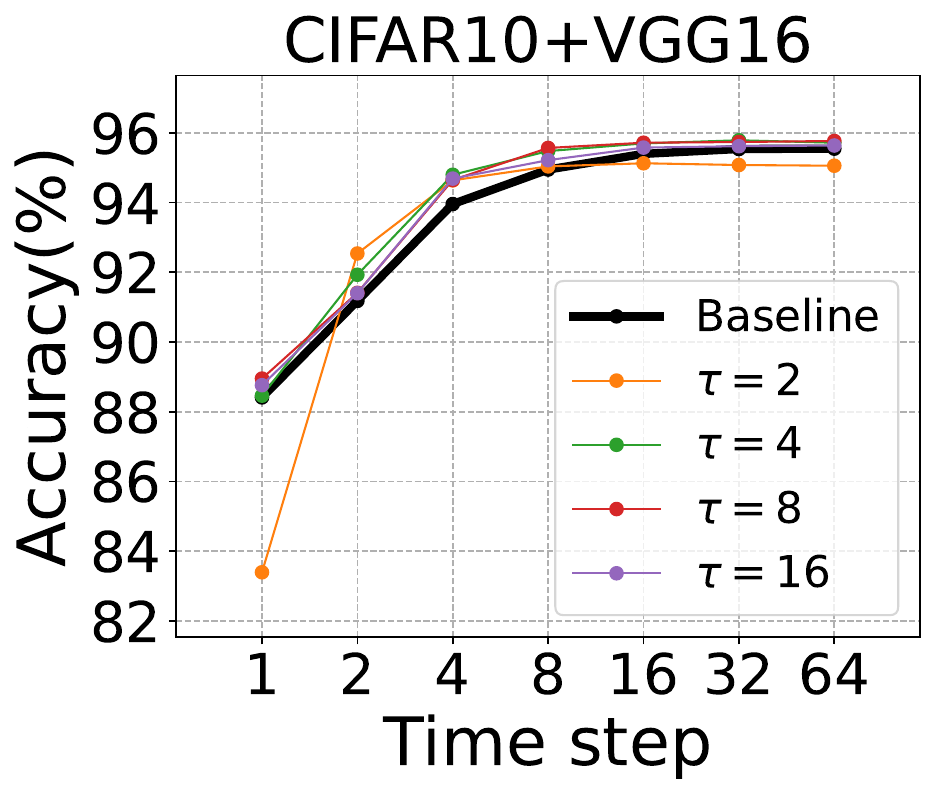}
  \end{minipage}%
  \begin{minipage}[t]{0.5\linewidth}
    \centering
    \includegraphics[scale=0.24]{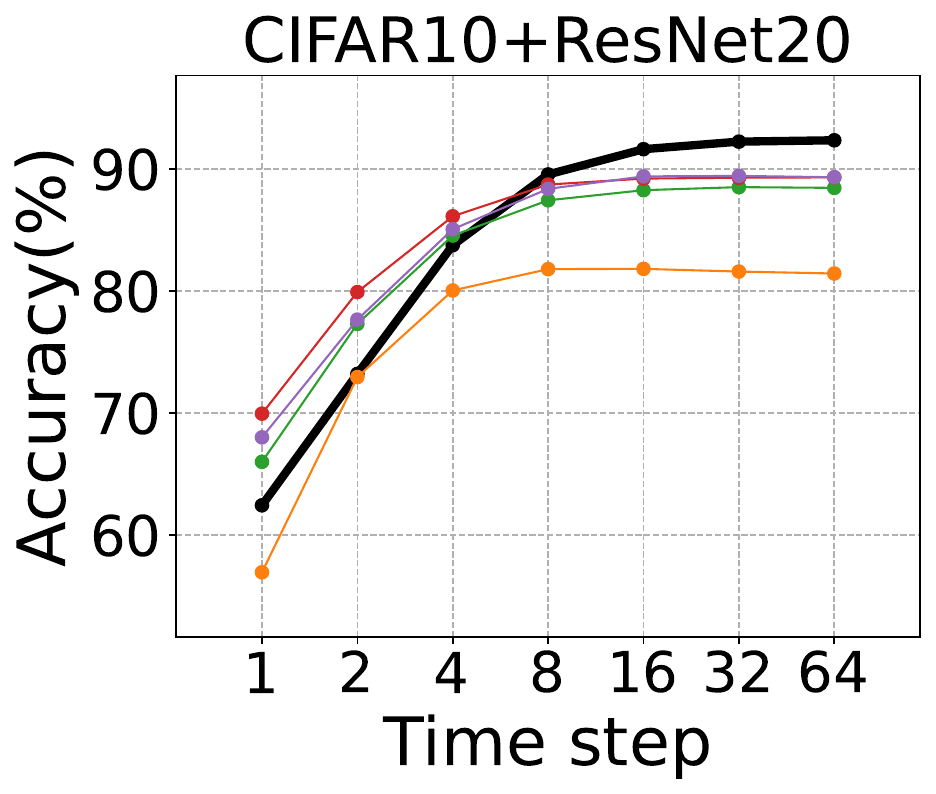}
  \end{minipage}
  \begin{minipage}[t]{0.5\linewidth}
    \centering
    \includegraphics[scale=0.24]{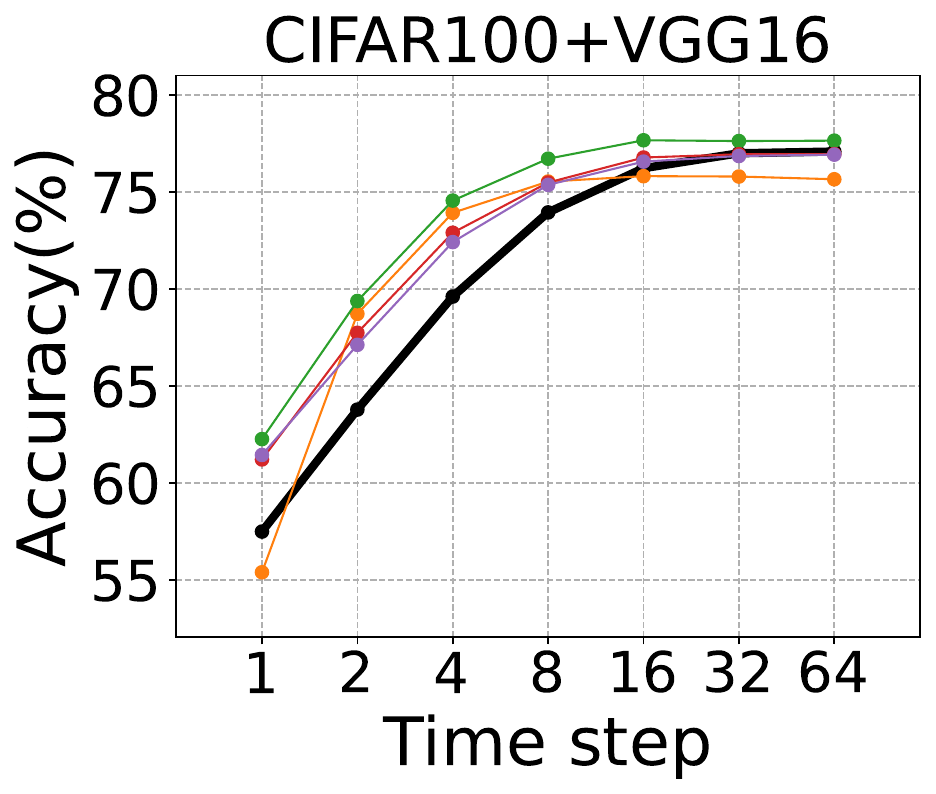}
  \end{minipage}%
  \begin{minipage}[t]{0.5\linewidth}
    \centering
    \includegraphics[scale=0.24]{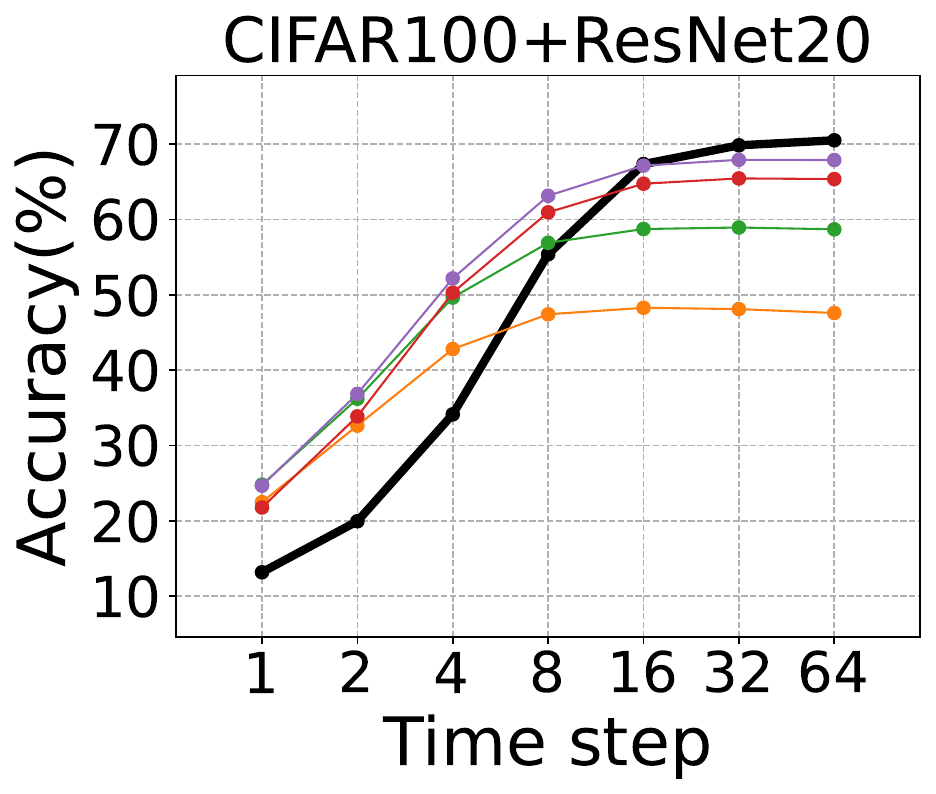}
  \end{minipage}
  \caption{Effect of noise-induction time step $\tau$}
  \label{fig:Validation time step}
\end{figure}

\subsection{Comparison With the State-of-the-Art Conversion Methods}

\begin{table*}[!t]
\caption{Comparison with existing state-of-the-art ANN-SNN conversion methods}
\label{tab:sota}
\small
\centering
\begin{tabular}{cccccccccc}
\hline
{Method}  & {ANN}     & {Architecture} & {T=1}     & {T=2}     & {T=4}     & {T=8}     & {T=16}    & {T=32}    & {T=64}    \\ \hline
\multicolumn{10}{c}{\textbf{CIFAR-10 Dataset}} \\ \hline
{SNM~\shortcite{wang2022signed}}     & {94.09\%} &                                     & {—}       & {—}       & {—}       & {—}       & {—}       & {93.43\%} & {94.07\%} \\ \cline{1-2} \cline{4-10} 
{RMP~\shortcite{han2020rmp}}     & {93.96\%} &                                     & {—}       & {—}       & {—}       & {—}       & {—}       & {60.30\%} & {90.35\%} \\ \cline{1-2} \cline{4-10} 
{RTS~\shortcite{deng2021optimal}}     & {95.72\%} &                                     & {—}       & {—}       & {—}       & {—}       & {—}       & {76.24\%} & {90.64\%} \\ \cline{1-2} \cline{4-10} 
{TCL~\shortcite{ho2021tcl}}     & {94.57\%} &                                     & {—}       & {—}       & {—}       & {—}       & {—}       & {93.64\%} & {94.26\%} \\ \cline{1-2} \cline{4-10} 
{RNL~\shortcite{ding2021optimal}}     & {92.82\%} &                                     & {—}       & {—}       & {—}       & {—}       & {57.90\%} & {85.40\%} & {91.15\%} \\ \cline{1-2} \cline{4-10} 
{SNNC-AP~\shortcite{li2021free}} & {95.72\%} &                                     & {—}       & {—}       & {—}       & {—}       & {—}       & {93.71\%} & {95.14\%} \\ \cline{1-2} \cline{4-10} 
{OPI~\shortcite{bu2022optimized}}     & {94.57\%} &                                     & {—}       & {—}       & {—}       & {90.96\%} & {93.38\%} & {94.20\%} & {94.45\%} \\ \cline{1-2} \cline{4-10} 
{QCFS~\shortcite{bu2023optimal}}    & {95.52\%} &                                     & {88.41\%} & {91.18\%} & {93.96\%} & {94.95\%} & {95.40\%} & {95.54\%} & {95.55\%} \\ \cline{1-2} \cline{4-10} 
{\textbf{Ours}}     & {95.21\%}                              & \multirow{-9}{*}{VGG16}     &{\textbf{88.46}\%}  &{\textbf{91.93}\%} &{\textbf{94.80}\%}         &{\textbf{95.48}\%}     &{95.70\%}         &{95.79\%}         &{95.72\%} \\ \hline

{RTS~\shortcite{deng2021optimal}}     & {92.32\%} &                                     & {—}       & {—}       & {—}       & {—}       & {92.41\%} & {93.30\%} & {93.55\%} \\ \cline{1-2} \cline{4-10} 
{SNM~\shortcite{deng2021optimal}}     & {95.39\%} &                                     & {—}       & {—}       & {—}       & {—}       & {—}       & {94.03\%} & {94.03\%} \\ \cline{1-2} \cline{4-10} 
{SNNC-AP~\shortcite{li2021free}} & {95.46\%} &                                     & {—}       & {—}       & {—}       & {—}       & {—}       & {94.78\%} & {95.30\%} \\ \cline{1-2} \cline{4-10} 
{OPI~\shortcite{bu2022optimized}}     & {96.04\%} &                                     & {—}       & {—}       & {—}       & {75.44\%} & {90.43\%} & {94.82\%} & {95.92\%} \\ \cline{1-2} \cline{4-10} 
{QCFS~\shortcite{bu2023optimal}}    & {96.04\%} &                                     & {—}       & {75.44\%} & {90.43\%} & {94.82\%} & {95.92\%} & {96.08\%} & {96.06\%} \\ \cline{1-2} \cline{4-10} 
{\textbf{Ours}}     &{95.52\%}  & \multirow{-6}{*}{ResNet18}       &{\textbf{89.64}\%}  &{\textbf{93.72}\%} &{\textbf{95.37}\%}      &{\textbf{96.21}\%} &{96.38\%} &{96.31\%}      &{96.36\%}        \\ \hline

{OPI~\shortcite{bu2022optimized}}     & {92.74\%} &                                     & {—}       & {—}       & {—}       & {66.24\%} & {87.22\%} & {91.88\%} & {92.57\%} \\ \cline{1-2} \cline{4-10} 
{QCFS~\shortcite{bu2023optimal}}    & {91.77\%} &                                     & {62.43\%} & {73.20\%} & {83.75\%} & {89.55\%} & {91.62\%} & {92.24\%} & {92.35\%} \\ \cline{1-2} \cline{4-10} 
{\textbf{Ours}}     & {85.18\%}  &   \multirow{-3}{*}{ResNet20}   &{\textbf{65.99}\%}  &{\textbf{77.30}\%} &{\textbf{84.54}\%}      &{87.43\%} &{88.26\%} &{88.51\%}      &{88.45\%}  \\ \hline

\multicolumn{10}{c}{\textbf{CIFAR-100 Dataset}}      \\ \hline
{SNM~\shortcite{deng2021optimal}}     & {74.13\%} &                                     & {—}       & {—}       & {—}       & {—}       & {—}       & {71.80\%} & {73.69\%} \\ \cline{1-2} \cline{4-10} 
{RTS~\shortcite{deng2021optimal}}     & {77.89\%} &                                     & {—}       & {—}       & {—}       & {—}       & {65.94\%} & {69.80\%} & {70.35\%} \\ \cline{1-2} \cline{4-10} 
{TCL~\shortcite{ho2021tcl}}     & {76.32\%} &                                     & {—}       & {—}       & {—}       & {—}       & {—}       & {52.30\%} & {71.17\%} \\ \cline{1-2} \cline{4-10} 
{SNNC-AP~\shortcite{li2021free}} & {77.89\%} &                                     & {—}       & {—}       & {—}       & {—}       & {—}       & {73.55\%} & {76.64\%} \\ \cline{1-2} \cline{4-10} 
{OPI~\shortcite{bu2022optimized}}     & {76.31\%} &                                     & {—}       & {—}       & {—}       & {60.49\%} & {70.72\%} & {74.82\%} & {75.97\%} \\ \cline{1-2} \cline{4-10} 
{QCFS~\shortcite{bu2023optimal}}    & {76.28\%} &                                     & {—}       & {63.79\%} & {69.62\%} & {73.96\%} & {76.24\%} & {77.01\%} & {77.10\%} \\ \cline{1-2} \cline{4-10} 
{\textbf{Ours}}     & {74.86\%}        & \multirow{-7}{*}{VGG16}        & {\textbf{62.27}\%}        & {\textbf{69.39}\%}   & {\textbf{74.57}\%}        & {\textbf{76.73}\%}        & {77.68\%}        & {77.64\%}        & {77.66\%}        \\ \hline

{RTS~\shortcite{deng2021optimal}}     & {67.08\%} &                                     & {—}       & {—}       & {—}       & {—}       & {63.73\%} & {68.40\%} & {69.27\%} \\ \cline{1-2} \cline{4-10} 
{SNNC-AP~\shortcite{li2021free}} & {77.16\%} &                                     & {—}       & {—}       & {—}       & {—}       & {—}       & {76.32\%} & {77.29\%} \\ \cline{1-2} \cline{4-10} 
{QCFS~\shortcite{bu2023optimal}}    & {78.80\%} &                                     & {—}       & {70.79\%} & {75.67\%} & {78.48\%} & {79.48\%} & {79.62\%} & {79.54\%} \\ \cline{1-2} \cline{4-10} 
{\textbf{Ours}}     & {76.66\%}        & \multirow{-4}{*}{ResNet18}          & {\textbf{62.35}\%}        & {\textbf{70.86}\%}        & {\textbf{76.81}\%}        & {\textbf{78.85}\%}        & {79.44\%}        & {79.62\%}        & {79.46\%}   \\ \hline

{RMP~\shortcite{han2020rmp}}     & {68.72\%} &                                     & {—}       & {—}       & {—}       & {—}       & {—}       & {27.64\%} & {46.91\%} \\ \cline{1-2} \cline{4-10} 
{OPI~\shortcite{bu2022optimized}}     & {70.43\%} &                                     & {—}       & {—}       & {—}       & {23.09\%} & {52.34\%} & {67.18\%} & {69.96\%} \\ \cline{1-2} \cline{4-10} 
{QCFS~\shortcite{bu2023optimal}}    & {69.94\%} &                                     & {—}       & {19.96\%} & {34.14\%} & {55.37\%} & {67.33\%} & {69.82\%} & {70.49\%} \\ \cline{1-2} \cline{4-10} 
{\textbf{Ours}}    & {62.34\%} &   \multirow{-4}{*}{ResNet20}                             & {\textbf{21.78}\%}       & {\textbf{33.87}\%} & {\textbf{50.28}\%} & {\textbf{60.93}\%} & {64.73\%} & {65.43\%} & {65.35\%} \\ \hline
\end{tabular}
\end{table*}

We compare our method with the state-of-the-art ANN-SNN conversion methods, including RMP~\cite{han2020rmp}, RTS~\cite{deng2021optimal}, RNL~\cite{ding2021optimal}, OPI~\cite{bu2022optimized}, SNNC-AP~\cite{li2021free}, TCL~\cite{ho2021tcl}, QCFS~\cite{bu2023optimal} and SNM~\cite{wang2022signed}.
Table~\ref{tab:sota} shows the performance of our proposed method. Our model almost outperforms all the other conversion methods when $2\leq T \leq 8$. On the CIFAR-10 dataset, for VGG-16, we achieve an accuracy 
of 94.80\%, very close to the accuracy of its ANN counterpart (95.21\%) with only 4 time steps. For ResNet-18, the accuracy of the converted SNN is 18.28\% higher than the current SOTA QCFS SNN at $T = 2$ and 4.94\% higher than the QCFS SNN at $T = 4$. Besides, on CIFAR-100, the proposed method achieves 60.93\% top-1 accuracy at 8 time steps for ResNet-20, which is 37.84\% and 5.56\% higher than OPI and QCFS, respectively. For VGG-16, we achieve 74.57\% top-1 accuracy when $T = 4$, which is 4.95\% higher than QCFS (69.62\%) at the same time step and 14.08\% higher than OPI at $T = 8$ (60.49\%).


\subsection{Comparison With Other SNN Training Methods}

Table~\ref{tab:Comparison With Other SNN Training Methods} presents the results between our proposed method and recent advanced SNN training methods on the CIFAR10/100 dataset, involving STBP-tdBN~\cite{zheng2021going}, TET~\cite{deng2022temporal}, TEBN~\cite{duan2022temporal}, Diet-SNN~\cite{rathi2021diet}, Dual-Phase ~\cite{wang2205towards}, RecDis-SNN~\cite{guo2022recdis} under the low latency condition. The accuracy achieved by our method when $T=4$ exceeds the accuracy of other SNN training methods using the same or even longer time step. It is worth emphasizing that our proposed method does not need to consume as much memory and computational resources as other SNN training methods to achieve high performance with very low inference latency. In addition, our method does not require any other additional operations or optimizations on the converted SNNs. This advantage enhances the usefulness of ANN-SNN conversion algorithms in neuromorphic chips.

\begin{table}[ht]
\caption{Comparison with other SNN training methods}
\label{tab:Comparison With Other SNN Training Methods}
\small
\centering
\begin{tabular}{ccccc}
\hline
{Method}     & {Type}    & {Arch} & {Acc} & {T} \\ \hline

\multicolumn{5}{c}{\textbf{CIFAR-10 Dataset}}   \\ \hline
{STBP-tdBN}  & {BPTT}    & {ResNet19}     & {93.16}    & {6}         \\ \hline
{TET}        & {BPTT}    & {ResNet19}     & {94.50}    & {6}         \\ \hline
{TEBN}       & {BPTT}    & {ResNet19}     & {94.71}    & {6}         \\ \hline
{\textbf{Ours}}       & {ANN2SNN} & {ResNet18}     & {95.37}    & {4}         
               \\ \hline
               
\multicolumn{5}{c}{\textbf{CIFAR-100 Dataset}}  \\ \hline
{Diet-SNN}   & {Hybrid}  & {VGG16}        & {69.67}   & {5}            \\ \hline
{Dual-Phase} & {Hybrid}  & {VGG16}      & {70.08}   & {4}            \\ \hline
{RecDis-SNN} & {BPTT}    & {VGG16}        & {69.88}   & {5}            \\ \hline
{\textbf{Ours}}       & {ANN2SNN} & {VGG16}        & {74.57}    & {4}         
               \\ \hline
\end{tabular}
\end{table}

\subsection{Effect of Noise-Induction Time Step} 
We further explored the effect of the noise-induction time step $\tau$. Fig.~\ref{fig:Validation time step} shows the accuracy of VGG16 and ResNet20 on CIFAR-10 and CIFAR-100 for different $\tau$ values. The settings of $L$ are the same as those in the experimental setup. When $\tau = L$, the noise intensity $delta$ mainly represents the effect of residual error during the validation process. 
When $\tau \ne L$, $delta$ represent the effect of both quantization error from $\tau \ne L$ and residual error from $\tau = L$.
We find that when $\tau=L$, for VGG16, SNN achieves better performance than the baseline at almost all time steps. For ResNet20, when $\tau = L$, the accuracy of SNN outperforms baseline at shorter time steps ($T\le \tau$). However, when $T>\tau$, the accuracy of SNN falls short of baseline. In short, the setting of $\tau=L$ can estimate the residual error more accurately, thus improving the performance of SNN with low latency ($T\le L$). 
The training process may primarily eliminate the performance gap of low-latency conversion, potentially introducing more noise and affecting ANN and thus SNN performance when $T\le\tau$.
Notably, this degradation problem can be solved by setting a larger $\tau$. The results in Fig.~\ref{fig:Validation time step} (purple lines) show that in the case of $\tau \ge L$, our method maintains a good generalization ability. 

\subsection{Training Overhead}

To assess the possible computational overhead associated with the inclusion of the validation process, we calculate the average training time per epoch of the CIFAR-100 dataset during the training of VGG16 and display the results in Fig.~\ref{fig:training_time}. 
The noise induction operation introduced to automatically adjust the noise intensity does not lead to a significant additional computational overhead, which shows that our training method is efficient and has potential practical applications.

\section{Conclusions}

This paper improves the low-latency ANN-SNN conversion by explicitly modeling an additive noise representing the effect of residual error in source ANNs, allowing the output of quantized activation to be closer to the average postsynaptic potential of spiking neurons. 
This approach maintains the expectation of conversion error, resulting in high accuracy and ultra-low latency. The method also proposes to adjust the noise intensity with low overhead, further reducing the performance gap between ANNs and SNNs under low-latency conditions. 
We evaluate the proposed method on the CIFAR10/100 dataset and showed that our method improves performance under low-latency conditions.
We believe that this work will facilitate the further development of neuromorphic hardware applications.

\bibliographystyle{named}
\bibliography{ijcai24}

\end{document}